\documentclass[final]{cvpr}

\usepackage{times}
\usepackage{epsfig}
\usepackage{graphicx}
\usepackage{amsmath}
\usepackage{amssymb}

\usepackage{threeparttable}
\usepackage{multirow}
\usepackage{array}
\usepackage{enumitem}
\usepackage{color}
\usepackage{booktabs}
\usepackage{amsmath}
\usepackage{pifont}
\usepackage{subfig}
\usepackage{color}

\usepackage[pagebackref=true,breaklinks=true,colorlinks,bookmarks=false]{hyperref}

\def\cvprPaperID{3851} 
\def\confYear{CVPR 2021}

\begin{document}

\title{OPANAS: One-Shot Path Aggregation Network Architecture Search for Object Detection}

\author{
	Tingting Liang$^{1}$\quad
	Yongtao Wang$^{1}$\thanks{indicates corresponding author.}\quad
	Zhi Tang$^{1}$\quad
	Guosheng Hu$^{2}$\quad
	Haibin Ling$^{3}$
	\\
	\vspace{-1em}
	\and
	$^{1}$Wangxuan Institute of Computer Technology, Peking University\quad
	$^{2}$Anyvision\\
	$^{3}$Department of Computer Science, Stony Brook University\\
	\vspace{-1em}
	{\tt\small tingtingliang, wyt, tangzhi@pku.edu.cn\quad huguosheng100@gmail.com\quad hling@cs.stonybrook.edu}\\
}

\maketitle

\thispagestyle{empty}
\begin{abstract}
Recently, neural architecture search (NAS) has been exploited to design feature pyramid networks (FPNs) and achieved promising results for visual object detection. Encouraged by the success, we propose a novel \emph{One-Shot Path Aggregation Network Architecture Search} (OPANAS) algorithm, which significantly improves both searching efficiency and detection accuracy. Specifically, we first introduce six heterogeneous information paths to build our search space, namely \emph{top-down}, \emph{bottom-up}, \emph{fusing-splitting}, \emph{scale-equalizing}, \emph{skip-connect} and \emph{none}. Second, we propose a novel search space of FPNs, in which each FPN candidate is represented by a densely-connected directed acyclic graph (each node is a feature pyramid and each edge is one of the six heterogeneous information paths). Third, we propose an efficient one-shot search method to find the optimal path aggregation architecture, that is, we first train a super-net and then find the optimal candidate with an evolutionary algorithm. Experimental results demonstrate the efficacy of the proposed OPANAS for object detection: (1) OPANAS is more efficient than state-of-the-art methods (\eg, NAS-FPN and Auto-FPN), at significantly smaller searching cost (\eg, only 4 GPU days on MS-COCO); (2) the optimal architecture found by OPANAS significantly improves main-stream detectors including RetinaNet, Faster R-CNN and Cascade R-CNN, by 2.3$\sim$3.2 $\%$ mAP comparing to their FPN counterparts; and (3) a new state-of-the-art accuracy-speed trade-off (52.2 $\%$ mAP at 7.6 FPS) at smaller training costs than comparable state-of-the-arts. Code will be released at \url{https://github.com/VDIGPKU/OPANAS.}

\end{abstract}
\section{Introduction}
	\label{introduction}
\begin{figure}[!t]
	\centering
	\includegraphics[width=1.0\linewidth]{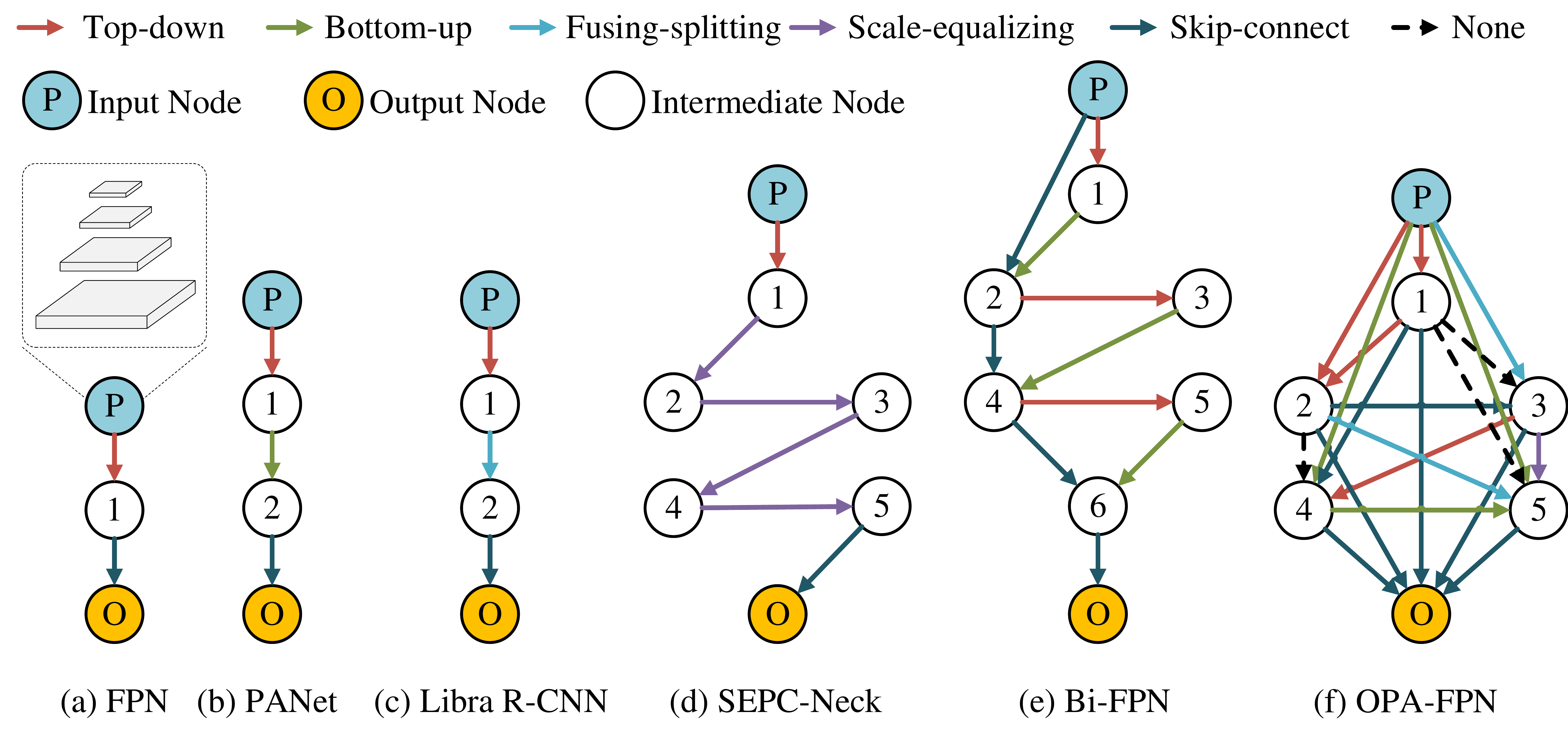}
	\caption{
		Different FPN architectures: (a) FPN \cite{LinDGHHB17}, (b) PANet \cite{LiuQQSJ18}, (c) Libra R-CNN \cite{DBLP:conf/cvpr/PangCSFOL19}, (d) SEPC-Neck \cite{DBLP:conf/cvpr/WangZYFZ20}, (e) BiFPN \cite{DBLP:journals/corr/abs-1911-09070}, and (f) our searched optimal FPN.}
	\label{fig:shouye}
	\vspace{-1em}
\end{figure}

\begin{figure*}[!t]
	\centering
	\includegraphics[width=1.0\linewidth]{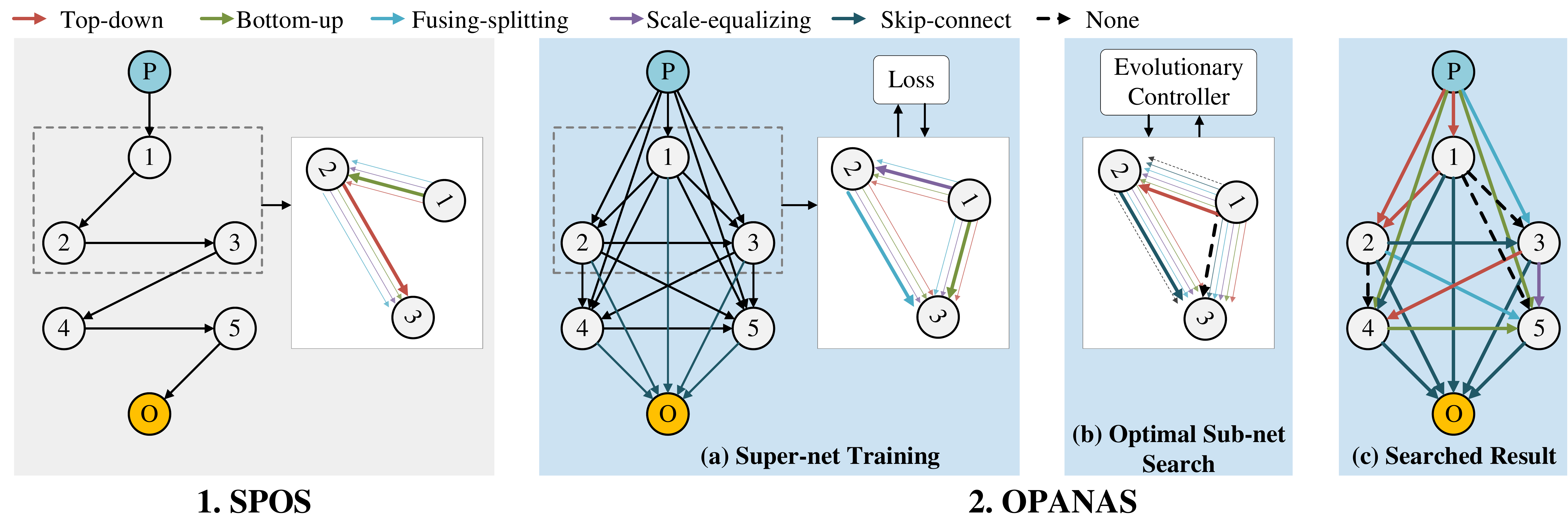}
	\caption{
		\textbf{1.} Single-path FPN super-net from SPOS search space \cite{DBLP:journals/corr/abs-1904-00420}. \textbf{2.} Our OPANAS:
		(a) super-net training, \textit{i.e.},  the optimization of super-net weights; 
		(b) optimal sub-net search with an evolutionary algorithm; 
		(c) the searched optimal architecture. 
		Note that two information paths (skip-connect and none) work only for (b).}
	\label{fig:super-net}
\vspace{-1em}
\end{figure*}

Recognizing objects at vastly different scales is one of the major challenges in computer vision. 
To address this issue, great progress has been made in designing deep convolutional networks in the past few years.
Intuitively, directly extracting feature pyramid \cite{LiuAESRFB16} from CNN at different stages provides an efficient solution. Each level of the feature pyramid corresponds to a specific scale in the original image. However, high-level features are with more semantics while the low-level ones are more content descriptive \cite{DBLP:conf/eccv/ZeilerF14}. Such a semantic gap is unable to deliver strong features for multi-scale visual recognition tasks (\textit{e.g.}, object detection, and segmentation).
To alleviate the discrepancy, different feature fusion strategies have been proposed. Feature Pyramid Network (FPN) \cite{LinDGHHB17} is arguably the most popular basic architecture and inspires many important variants. It adopts a backbone model, typically designed for image classification, and builds a top-down information flow by sequentially combining two adjacent layers in feature hierarchy in the backbone. 
By such design, low-level features are complemented by semantic information from high-level features. Despite simple and effective, FPN may not be the optimal architecture design. 

Two lines of research have been conducted to advance FPN-based detection algorithms. On one hand, various approaches (\textit{e.g.}, PANet  \cite{LiuQQSJ18}, BiFPN  \cite{DBLP:journals/corr/abs-1911-09070}, Libra R-CNN \cite{DBLP:conf/cvpr/PangCSFOL19} and SEPC \cite{DBLP:conf/cvpr/WangZYFZ20}) enrich FPN by aggregating multiple heterogeneous information paths and achieve impressive results. However, as shown in Fig.~\ref{fig:shouye} (a-e), {they only explore aggregations of up to \textit{three} types of information paths}, (\ie, \textit{top-down and bottom-up}~\cite{LiuQQSJ18},  \textit{top-down and fusing-splitting}~\cite{DBLP:conf/cvpr/PangCSFOL19}, and \textit{top-down and scale-equalizing}~\cite{DBLP:conf/cvpr/WangZYFZ20}). Moreover, most of these methods follow a straightforward \textit{chain-style aggregation structure}, 
except BiFPN that adds additional skip-connect on PANet with several repetitions, but \textit{remains in a  simple topology}.
On the other hand, Neural Architecture Search (NAS)-based FPN architectures~\cite{DBLP:conf/cvpr/GhiasiLL19, DBLP:conf/cvpr/WangGCWTSZ20, Xu_2019_ICCV} have achieved remarkable performance gain beyond manually designed architectures, but with following limitations: (1) \textit{inefficiency}, the searching processes are often computationally expensive (\textit{e.g.}, 300 TPU days~\cite{DBLP:conf/cvpr/GhiasiLL19}) due to the extremely large search space, and (2) \textit{weak adaptability}, their searched  architectures are specialized for certain detector with special training skills (\eg, large batch size or longer training schedule).

\begin{figure*}[!t]
	\centering
	\includegraphics[width=1.0\linewidth]{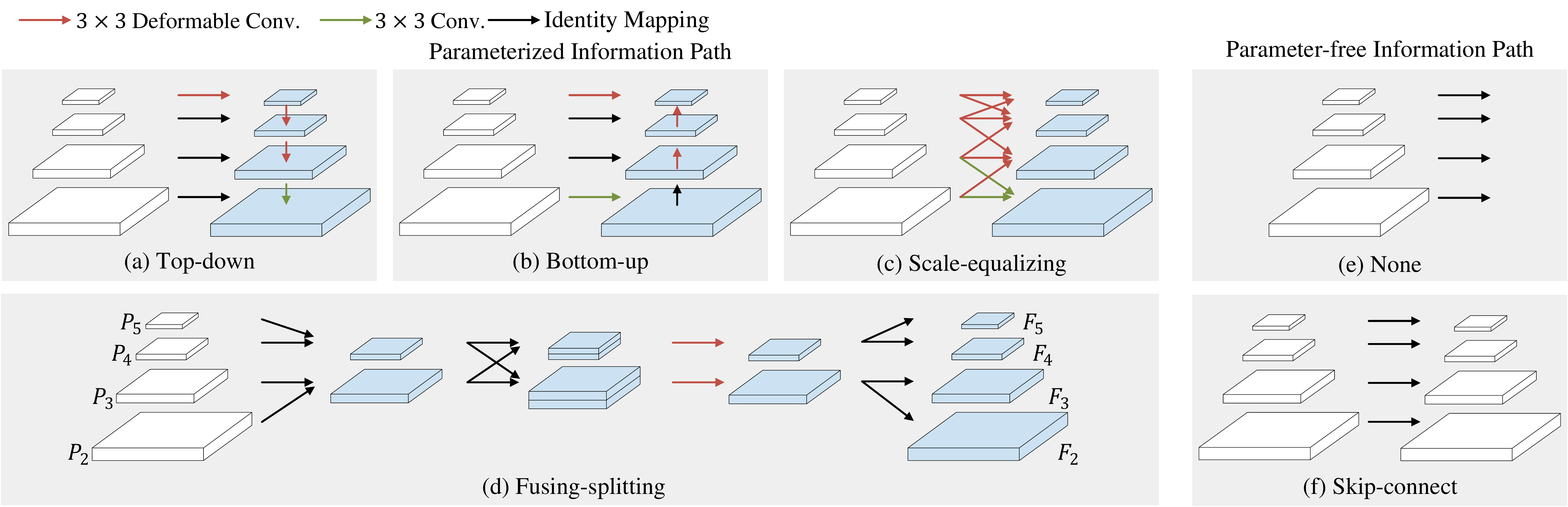}
	\caption{The proposed six heterogeneous information paths mapping 4-level pyramid features $\{P_2, P_3, P_4, P_5\}$ to $\{F_2, F_3, F_4, F_5\}$. (a)-(d) are parameterized and (e)-(f) are parameter-free.	}
	\label{fig:searchspace}
	\vspace{-1em}
\end{figure*}

Inspired by these studies and meanwhile to address aforementioned issues, we propose a new efficient and effective NAS framework, named OPANAS (One-Shot Path Aggregation Neural Architecture Search, see Fig.~\ref{fig:super-net}) to automatically search a better FPN for object detection. 
Firstly, we carefully design four parameterized information paths (\textit{top-down}, \textit{bottom-up}, \textit{scale-equalizing} and \textit{fusing-splitting}, see Fig.~\ref{fig:searchspace} (a-d)) and two parameter-free ones (\textit{skip-connect} and \textit{none}, see Fig.~\ref{fig:searchspace} (e-f)) to build our search space. Clearly, these six modules introduce different information flows, different connections between backbone and detection head, and lead to complementary and highly interpretable aggregation modules. Note that the four parameterized ones are relatively heavy and the two parameter-free ones are light-weighted, and they work together to achieve a promising accuracy-efficiency trade-off. 

Secondly, to achieve the optimal aggregation of the six information paths, we propose a novel FPN search space, in which each FPN candidate is represented by a densely-connected directed acyclic graph (each node is a feature pyramid and each edge is a specific one of the six heterogeneous information paths as shown in Fig.~\ref{fig:searchspace}). Notably, our search space contains richer aggregation topological structures of FPNs than existing methods as in Fig.~\ref{fig:shouye}, and hence enables richer cross-level and cross-module interactions. 

Thirdly, we propose an efficient one-shot search method to search the optimal FPN architecture, that is, we first train a super-net and then search the optimal sub-net from the super-net with an evolutionary algorithm that has strong global optimum search capability. Experiments show that our method is efficient as the differentiable NAS methods, \ie, DARTS~\cite{DBLP:conf/iclr/LiuSY19} and Fair DARTS~\cite{DBLP:journals/corr/abs-1911-12126}, while the searched FPN architecture can achieve better detection accuracy with less parameters and FLOPs. Moreover, following the simple vanilla training protocol, our searched FPN architecture can consistently improve the detection accuracy of the main-stream detectors including RetinaNet, Faster R-CNN and Cascade R-CNN by 2.3$\sim$3.2 mAP, with less parameters and FLOPs. These results demonstrate the efficacy of the proposed OPANAS for object detection.

Our contributions can be summarized as:
\begin{itemize}
\vspace{-0.175cm}
	\item We carefully design 6 information paths that can aggregate multi-level information, and thus enable the effective and complementary combination of low-, medium- and high-level information. To our knowledge, we are the first to investigate the aggregations of multiple ($>$3) information paths.  
\vspace{-0.175cm}
	\item We propose a novel one-shot method, OPANAS,  to efficiently and effectively search the optimal aggregation of the 6 kinds of information paths.
\vspace{-0.175cm}
	\item Working as a plug and play module, our searched architecture can easily be  adapted to main-stream detectors including RetinaNet, Faster R-CNN and Cascade R-CNN, and significantly improve their detection accuracy by 2.3$\sim$3.2 $\%$ mAP. Notably, we achieve a new state-of-the-art accuracy-speed trade-off (52.2 $\%$ mAP at 7.6 FPS).
\end{itemize}

\section{Related Work}
\subsection{Object detection}
Existing deep learning-based detectors can be briefly categorized into two streams: one-stage detectors such as SSD~\cite{LiuAESRFB16} and RetinaNet~\cite{LinGGHD17}, which utilize CNN directly to predict the bounding boxes; and two-stage methods such as Faster R-CNN~\cite{RenHGS15} and Mask R-CNN~\cite{HeGDG17}, which generate the the final detection results after extracting region proposals upon a region proposal network (RPN). Although encouraging signs of progress have been made, existing detectors are still suffering from the problems caused by the scale variation across object instances.
The feature pyramid is popularly used to deal with scale variation~\cite{LinDGHHB17}, which introduces a top-down information flow.

Beyond FPN, some recent extensions employ two or three types of information paths. For example, PANet~\cite{LiuQQSJ18} introduces an extra bottom-up path after the top-down path of classic FPN~\cite{LinDGHHB17}, and Libra R-CNN~\cite{DBLP:conf/cvpr/PangCSFOL19} adopts Non-Local module~\cite{DBLP:conf/cvpr/0004GGH18} to fuse the features produced by the classic FPN~\cite{LinDGHHB17} and then transfers the fused feature into multi-scale pyramid features. Multi-level FPN~\cite{zhao2019m2det}
first fuses the backbone features as the base feature and then introduces multiple U-shape modules
to extract multi-level pyramid features and builds a powerful one-stage detector. SEPC~\cite{DBLP:conf/cvpr/WangZYFZ20} stacks 4 scale equalizing modules behind classic FPN to enhance cross-scale correlation.
More recently, BiFPN~\cite{DBLP:journals/corr/abs-1911-09070} exploits a simplified architecture of PANet and stacks it repeatedly with skip-connect to build a more powerful one-stage detector named EfficientDet. Though promising results are achieved by EfficientDet, its training cost is extremely expensive, \textit{i.e.}, large batch-size (128 on 32TPU) with a long training schedule (300 or 500 epochs). Generally speaking, these FPNs suffer from intrinsic architecture limitations since they only aggregate at most three types of information paths with naive topological structure. 

\subsection{Neural Architecture Search}
More recently, neural architecture search (NAS) is applied to automatically search an FPN architecture for a specific detector. NAS-FPN~\cite{DBLP:conf/cvpr/GhiasiLL19}, NAS-FCOS~\cite{DBLP:conf/cvpr/WangGCWTSZ20} and SpineNet~\cite{DBLP:conf/cvpr/DuLJGTCLS20} use reinforcement learning to control the architecture sampling and obtain promising results. SM-NAS~\cite{DBLP:journals/corr/abs-1911-09929} uses evolutionary algorithm and partial order pruning method to search the optimal combination of different parts of the detectors. The above NAS methods are effective though can be time-consuming. Auto-FPN~\cite{Xu_2019_ICCV}, Hit-Detector~\cite{DBLP:conf/cvpr/GuoHWZYW0020} uses gradient-based method to search the optimal detector, which can significantly reduce searching time. However, gradient-based methods tend to trap into local minima in certain nodes of super-net during the progress of optimization and introduce further complexity~\cite{DBLP:journals/corr/abs-1904-00420}.
Recently, researchers~\cite{DBLP:conf/icml/BenderKZVL18, DBLP:conf/iclr/BrockLRW18, DBLP:journals/corr/abs-1904-00420} propose one-shot method to decouple the super-net training and architecture search in two sequential steps. DetNAS~\cite{DBLP:conf/nips/ChenYZMXS19} follows this idea to search  an efficient backbone for object detection. One limitation of the single-path approach is that the search space is restricted to a sequential structure as shown in Fig. \ref{fig:super-net}.1.

The aforementioned methods take the layer-wise operations as transform blocks (\textit{i.e}, single-scale feature as nodes), which are completely separated from manual design. Such design forms a large search space which contains architectures beyond human design, while also includes many poor-performing architectures, leading to low search efficiency.
 To reduce the post-processing overhead, we propose multi-level information path aggregation as our search space. With the help of carefully designed information paths, our search can be efficient and robust. 
\section{Methodology}
In this work, we first propose six types of information paths, which capture diverse multi-level information. Second, to search the optimal aggregations of these information paths, we introduce an efficient One-Shot Path Aggregation Network Architecture Search (OPANAS) algorithm. Last, we detail the optimization and searching process. 

\subsection{Information Paths}
To effectively aggregate different levels of pyramidal features, we propose 6 information paths, which can capture low-, medium- and high-level information. Similar to classic FPN~\cite{LinDGHHB17}, these information paths map the input pyramidal features $\{P_2,P_3,P_4,P_5\}$ (see Fig.~\ref{fig:searchspace}) to $\{F_2,F_3,F_4,F_5\}$. However, the proposed information paths can capture much richer and diverse information than FPN, which will be described as following. 
\paragraph{Top-down Information Path}
The top-down information path is modified from the classic FPN \cite{LinDGHHB17} (Fig.~\ref{fig:searchspace} (a)). 
For this path, 
the output pyramidal features (denoted as $F_2^{t},F_3^{t},F_4^{t},F_5^{t}$) are sequentially constructed in a top-down manner, \textit{i.e.}, the smaller scale (high-level, \textit{e.g.}, $F_5^{t}$) feature map is constructed first. Specifically, each feature map ($F_i^{t}$) is iteratively built by combining input pyramid feature map of the same level ($P_i$) and the higher-level output feature ($F_{i+1}^{t}$):
\begin{equation}
	{F_i^{t}=\mathbf{W}^t_{i}\otimes (\textit{U}(F^t_{i+1})+P_{i}),}
\end{equation}
where $\textit{U} (\cdot)$ denotes upsampling with factor of 2. For high-level features $ (i=3,4,5)$, $\mathbf{W}_i^t$ is the $3\times3$ deformable convolution filter to alleviate discrepancy of a feature pyramid \cite{DBLP:conf/cvpr/WangZYFZ20}, and $\mathbf{W}_2^t$ is a normal $3\times3$ convolution filter.
\paragraph{Bottom-up Information Path}
For the bottom-up information path, the output pyramidal features (denoted as $F_2^{b},F_3^{b},F_4^{b},F_5^{b}$) are sequentially constructed in a bottom-up manner, \textit{i.e.}, the large scale (low-level, \textit{e.g.}, $F_2^{b}$) feature map is constructed first as shown in Fig.~\ref{fig:searchspace} (b). Each feature map ($F_i^{b}$) is obtained by merging the input feature maps ($P_{i}$) of the same level, and the output feature map below it ($F_{i-1}^{b}$): 
\begin{equation}
	{
		F_{i}^{b}=\mathbf{W}_{i}^b\otimes(\textit{D}(F^b_{i-1})+P_i),
	}
\end{equation}
where $\textit{D}(\cdot)$ denotes downsampling with factor of 2 and $\mathbf{W}_i^b$ is the convolution filter with the same setting as above. 

\paragraph{Scale-equalizing Information Path}
The scale-equalizing information path is motivated by SEPC~\cite{DBLP:conf/cvpr/WangZYFZ20}, 
which stacks scale-equalizing pyramid convolutions after the classic FPN to capture inter-scale correlation. Here we take a single pyramid convolution operation as an information path. As shown in Fig.~\ref{fig:searchspace} (c), each feature map ($F_i^{s}$) is obtained by merging the adjacent-level input feature maps ($P_{i}$): 
\begin{equation}
	{
		F_{i}^{s}=\textit{U}(\mathbf{W}_{1}^s\otimes P_{i+1}) +\mathbf{W}_{0}^s\otimes P_{i} + \mathbf{W}_{-1}^s\otimes P_{i-1} ,
	}
\end{equation}
where $\mathbf{W}^s_{1}, \mathbf{W}^s_{0}, \mathbf{W}^s_{-1}$ are $3\times3$ deformable convolution filters and the stride of $\mathbf{W}^s_{-1}$ is set to 2 to down-sample.
\paragraph{Fusing-splitting Information Path}
We design a two-step fusing-splitting information path,
which first combines the high- and low-level input pyramidal features, and then splits the combined features to multi-scale output features in Fig.~\ref{fig:searchspace} (d).
In practice, the highest two input feature maps are merged into $\alpha_s$, and the lowest two are merged into $\alpha_l$ through an element-wise sum:
\begin{equation}
	{
		\alpha_s=P_4+\textit{U}(P_{5}), \ \
		\alpha_l=\textit{D}(P_2)+P_{3}.
	}
\end{equation}
After obtaining the combined features, we fuse them through concatenation, 
\begin{equation}
	{	\begin{aligned}
		\beta_s=\mathbf{W}_{s}^f\otimes \mathrm{concat}(\alpha_s, \textit{D}(\alpha_l)),\\
		\beta_l=\mathbf{W}_{l}^f\otimes \mathrm{concat}(\textit{U}(\alpha_s), \alpha_l),
		\end{aligned}
	}
\end{equation}
where $\mathbf{W}_{s}^f$ and $\mathbf{W}_{l}^f$ are $3\times3$ deformable convolution filters, and $\mathrm{concat}(\cdot)$ represents concatenation along the channel dimension. 
After these operations, feature maps $\beta_s, \beta_l$ carry information fused from all level features. Finally, we resize them into multi-scale pyramid feature maps, 
\begin{equation}
	{
		F_2^{f}=\textit{U}(\beta_l),
		F_3^{f}=\beta_l;\\
		F_4^{f}=\beta_s,
		F_5^{f}=\textit{D}(\beta_s).
	}
\end{equation}
\paragraph{Skip-connect Information Path and None}
Specially, we add a skip-connect path to perform identity mapping. 
Moreover, a ‘none’ information path is exploited to remove redundant information paths. These two parameter-free information paths are designed to reduce the complexity of the model, leading to a better accuracy-efficiency trade-off.
\begin{table*}[t]
\vspace{-1mm}
\begin{centering}
\small
\tabcolsep 0.01in{\scriptsize{}}
\begin{threeparttable}
\begin{tabular}
{l|ccccccc}
\hline\hline
\multirow{2}{*}{\textbf{Method}} &\multirow{2}{*}{\textbf{Backbone}}&\multirow{2}{*}{\textbf{Time (fps)}}&\multirow{2}{*}{\textbf{FLOPs}}&\multirow{2}{*}{\textbf{Params}}&\multirow{2}{*}{\textbf{Search Part}}&\textbf{Search Cost}&\multirow{2}{*}{\textbf{mAP}}\\
&&&&&&(\textbf{GPU-day})&\\
\hline
NAS-FPN(7@ 256)\cite{DBLP:conf/cvpr/GhiasiLL19}&ResNet50&$17.8^{\dagger}$ &281G &60.3M  &Neck&333$\times \#$TPUs&39.9\\
DetNAS-FPN-Faster\cite{DBLP:conf/nips/ChenYZMXS19}&Searched&-  &- &- &Backbone&44&40.2\\
Auto-FPN\cite{Xu_2019_ICCV} &ResNet50&7.7 &260G &32.5M  &Neck $\&$ Head&16&40.5\\
\textbf{Faster OPA-FPN@64}&ResNet50 &\textbf{22.0}  &\textbf{123G} &\textbf{29.5M} &Neck&\textbf{4}&\textbf{41.9}\\
\hline
Auto-FPN\cite{Xu_2019_ICCV} &X-64x4d-101&5.7 &493G &90.0M  &Neck $\&$ Head&16&44.3\\
SM-NAS\cite{DBLP:journals/corr/abs-1911-09929}&Searched&$9.3^{\dagger}$  &-&- &Backbone $\&$ Neck $\&$ Head&188&45.9\\
NAS-FCOS(@128-256)\cite{DBLP:conf/cvpr/WangGCWTSZ20} &X-64x4d-101&- &362G &-  &Neck $\&$ Head&28&46.1\\
\textbf{Cascade OPA-FPN@160}&ResNet50&\textbf{12.6}  &\textbf{326G} &\textbf{60.6M} &Neck&\textbf{4}&\textbf{47.0}\\
\hline
NAS-FPN(7@ 384)$\&$DropBlock\cite{DBLP:conf/cvpr/GhiasiLL19}&AmoebaNet&$3.6^{\dagger}$ &1126G &166.5M  &Neck&333$\times \#$TPUs&48.3\\
SP-NAS\cite{DBLP:conf/cvpr/JiangXZLL20}&Searched&$2.1^{\dagger}$  &949G&- &Backbone $\&$ Neck $\&$ Head&$>26$&49.1\\
EfficientDet-D7 ($1536\times 1536$)\cite{DBLP:journals/corr/abs-1911-09070}  &EfficientNet-B6  &$3.8^{\dagger}$&\textbf{325G} &\textbf{51.9M} &-&-
&\textbf{52.2}\\
SpineNet-190\cite{DBLP:conf/cvpr/DuLJGTCLS20}&SpineNet-190&- &1885G &164.0M &Backbone $\&$ Neck &700$\times \#$TPUs&52.1\\
\textbf{Cascade OPA-FPN@160} ($1200\times 900$)&Res2Net101-DCN &\textbf{7.6}  &432G &80.3M &Neck&\textbf{4}&\textbf{52.2}\\
\hline
\end{tabular}
 \begin{tablenotes}
       \footnotesize
       \item[1] $\dagger$ FPS marked with $\dagger$ are from papers, and all others are measured on the same machine with 1 V100 GPU. 
\end{tablenotes}
\end{threeparttable}
{\scriptsize\par}
\par\end{centering}	

\vspace{1mm}	
\caption{
Comparison with SOTA methods on COCO \texttt{test-dev} set . \textbf{Here and after, ‘@$c$’ denotes the feature channel is $c$, \textit{e.g.}, ‘@160’ implies that the feature channel is 160.}  }
\label{tab:comparenas}
\vspace{-1mm}

\end{table*}

\subsection{One-Shot Search}
We propose a one-shot search method to efficiently and effectively search the optimal aggregation of the above six types of information paths.
Specifically, we first construct a super-net $\mathcal{A}$, which is a fully-connected Multigraph  DAG (directed acyclic graph). 
The node of DAG stands for feature maps (in the way of a feature pyramid), and there are six edges of different types between two nodes, and each edge represents one information path. The whole optimization includes two steps: 
(i) super-net training and (ii) optimal sub-net search, as shown in  Fig.~\ref{fig:super-net}. For (i), we train the super-net until convergence (optimization of the weights of the super-net) using a fair sampling strategy detailed in Section~\ref{sec:fairsampling}. The weights of super-net are fixed once this training is done (one-shot optimization). For (ii), we use evolutionary algorithm (EA) to search for the optimal sub-net $a^{*}$, which is a DAG with only one optimal edge between two nodes. Obviously, the optimal sub-net represents the desired optimal FPN aggregating multiple information paths.  Note that (ii) is very efficient because each sampled sub-net $a$ just goes through the inference process by using the weights of the super-net trained in (i). This is the main reason why one-shot optimization is very efficient.

To detail the optimization process, we first introduce the components of the super-net. The super-net is a DAG consists of $N +2$ nodes ($N$ is a predefined constant value), where the input node $P$ represents the extracted feature from the backbone, and the output node $O$ is the final output feature pyramid. Similarly, intermediate nodes ${x}_{i} (i=1,2,...,N)$ are also feature pyramids. Each directed edge $(i, j)$ is associated with some information path $\mathbf{IP}(i, j)$ that transforms ${x}_{i}$ to $x_j$. We assume the intermediate nodes are fully connected with former nodes, and identity mapped to the output node through summation. In such DAG model, each node $i \in \{1,2,\ldots,N\}$ aggregates inputs from previous nodes, where 
\begin{equation}
x_j = \sum_{i<j} \textbf{IP}(i, j)(x_i).
\label{eq:transform}
\end{equation} 
In this way, OPANAS allows $6^{{N(N+1)}/2}$ possible DAGs without considering graph isomorphism with $N$ intermediate nodes. In particular, to maximize the search space without affecting the convergence, we set $N=5$, and the total number of sub-nets is approximately $6^{15} \approx 10^{12}$.

Second, we formulate the super-net training as following. 
The architecture space $\mathcal{A}$ is encoded in a super-net, denoted as $\mathcal{N}(\mathcal{A}, W)$, where $W$ stands for the weights of super-net. Thus, the super-net training can be formulated as: 
\begin{equation} W_{\mathcal{A}}=\underset{W}{\operatorname{argmin}} \mathcal{L}_{\text {train }}(\mathcal{N}(\mathcal{A}, W)),
\label{eq:op}
\end{equation}
This super-net training is detailed in Section \ref{sec:fairsampling}.

Third, we discuss the optimal sub-net search in (ii). We aim to search the optimal sub-net $a^{*} \in \mathcal{A}$ that maximizes the validation accuracy, which can be formulated as:
\begin{equation}
	a^{*}=\underset{a \in \mathcal{A}}{\operatorname{argmax}} \mathrm{ACC}_{\text {val }}\big(\mathcal{N}\left(a, W_{\mathcal{A}}(a)\right)\big).
\end{equation}
We use an evolutionary algorithm to conduct this optimal sub-net search detailed in Section \ref{sec:subnet search}.

\subsubsection{Super-net Training}
\label{sec:fairsampling}

\paragraph{Edge Importance Weighting}
Unlike the existing One-Shot method SPOS \cite{DBLP:journals/corr/abs-1904-00420} only having edges between adjacent nodes, our OPANAS is densely connected to explore richer topological structures for aggregation. To adapt to our multiple paths (edges) optimization, we associate an \textit{edge importance weight} to each edge.
To guarantee the consistency between training and test, we set these weights to be continuous. Consequently, each node $i \in \{1,2,\ldots,N\}$ aggregates weighted inputs from the previous nodes, then Eq.~(\ref{eq:transform}) can be formulated as: 
\begin{equation}
	x_j = \sum_{i<j} \gamma_{i, j} \textbf{IP}(i, j)(x_i),
\label{eq:gamma}
\end{equation}
where $\gamma_{i, j}$ denotes the edge importance weight  between node $i$ and $j$. Concomitantly, the optimization in Eq.~(\ref{eq:op}) is modified as:
\begin{equation}
W_{\mathcal{A}}=\underset{W, \gamma}{\arg \min } ~ \mathcal{L}_{\text {train }}(\mathcal{N}(\mathcal{A}, W, \gamma)).
\label{eq:mpos}
\end{equation}
To assist the convergence of model, 
we add $L_1$ regularization to these edge importance weights with a hyper-parameter $\mu$ to balance with the original bounding box loss. Thus the total loss function is:
\begin{equation}
	\begin{aligned}
	\mathcal{L}&=\mathcal{L}_{\rm bbox}+\mu \mathcal{L}_{1} \\
&=\sum{(\mathcal{L}_{\rm cls}+ \mathcal{L}_{\rm loc})} +\mu \|\gamma\|_{1}.
	\end{aligned}
\end{equation}
$\mathcal{L}_{\rm cls}, \mathcal{L}_{\rm loc}$ are objective functions corresponding to recognition and localization task respectively.

\paragraph{Fair Sampling}
Instead of training the whole super-net directly, we sample $K$ sub-nets per training iteration to reduce the GPU memory cost. Note that `skip-connection' and `none' are parameter-free and do not require any optimization. Hence, they are only considered during the searching process. Consequently, in super-net training, only $K=4$ types of information paths are involved. 
To alleviate training unfairness between the $K$ parameterized information paths, we adopt strict fair sampling strategy \cite{DBLP:journals/corr/abs-1907-01845} in our super-net training. To be more specific, in the $n$-th super-net training step, $K$ sub-nets are sampled with no intersection. That is, each edge of them is associated with different information path, and the weights of the super-net  are updated after accumulating gradients from the $K$ sampled sub-nets. By this sampling strategy, all information paths are ensured to be equally sampled and trained within each training step, and each edge is activated only once within each training step. Consequently, the expectation and variance of edge $E_{i}$ with information path $i$ ($i=0,1,2$ and $3$  correspond to \textit{top-down}, \textit{bottom-up}, \textit{scale-equalizing} and \textit{fusing-splitting}, respectively) are given by,
\begin{equation}
	{
		\mathbf{E}(Y_{E_i})=n \times P_{E_i}=n/K,\ \		
		\mathbf{Var}(Y_{E_i})=0.
	}
\end{equation}
The variance does not change with $n$, thus fairness is assured at every training step. 

\subsubsection{Sub-net Search with Evolutionary Algorithm}
We conduct the sub-net search with an evolutionary algorithm. 
Specifically, during the optimal sub-net search in Fig.~\ref{fig:super-net} (b), we first randomly sample $N_S$ sub-nets, each passes the coarse search, from the super-net and rank their performance. Note that evaluating a sub-net requires only inference without training, which makes the search very efficient. Then we repeatedly generate new sub-nets through crossover and mutation on top $k$ performing sub-nets. Following an evolutionary algorithm \cite{DBLP:journals/corr/abs-1904-00420}, crossover denotes that two randomly selected sub-nets are crossed to produce a new one, mutation means a randomly selected sub-net mutates its every edge with probability 0.1 to produce a new sub-net. In this work, we set population size $N_S = 50$, max iterations $T = 12$ and $k = 10$. 
\label{sec:subnet search}

\section{Experiments}
\begin{table}[t]
	\vspace{-1mm}
\begin{centering}
	\small
	\tabcolsep 0.05 in{\scriptsize{}}
	\begin{tabular}
	{l|cccccc}
		\hline\hline
		\textbf{Detector} &\textbf{Method }  &\textbf{FLOPs} &\textbf{Params} &\textbf{mAP} &\textbf{FPS}\\  
		\hline
		\multirow{2}{*}{\shortstack{RetinaNet}}
		&Baseline  &239G&37.7M & 35.7 &\textbf{19.7}\\
	    &SEPC-Neck &314G &45.3M  &\textbf{38.0} &15.1\\
		&Balanced FPN  &240G &38.0M  &36.4 &18.5\\
		&PAFPN &245G &40.1M  & 35.9 &17.8\\
		&\textbf{OPA-FPN@168}&\textbf{207G} &\textbf{36.5M} &$\mathbf{38.0}$ &18.1\\
		\hline
		\multirow{2}{*}{\shortstack{Faster\\ R-CNN }}
	&Baseline   &207G  &41.5M
	&36.4 &\textbf{20.6}\\
	
	&SEPC-Neck &509G &49.1M  & 39.0& 10.9\\
	&Balanced FPN  &208G &41.8M  & 37.2& 19.3\\
	&PAFPN   &232G &45.1M  & 36.5& 19.1\\
	&\textbf{OPA-FPN@112} &\textbf{197G} &\textbf{35.5M}   &$\mathbf{39.6}$ &17.3\\

\hline
		\multirow{2}{*}{\shortstack{Cascade  \\R-CNN }}
		&Baseline  &235G  &69.2M
		&40.3 &\textbf{18.1}\\
		&SEPC-Neck &536G &76.3M  &42.6 &9.9\\
		&Balanced FPN  &236G &69.4M &41.2 &17.0\\
		&PAFPN  &259G &72.7M  &40.5 &16.8\\
		&\textbf{OPA-FPN@120}  &\textbf{225G} &\textbf{50.6M}&$\mathbf{42.8}$ &15.0\\

		\hline
	\end{tabular}{\scriptsize\par}
\par\end{centering}		
\vspace{1mm}		
	\caption{
	Comparisons of model adaptability for main-stream detectors on COCO \texttt{minival} with FPN \cite{LinDGHHB17} (baseline), SEPC-Neck  \cite{DBLP:conf/cvpr/WangZYFZ20} (stacking 4 scale-equalizing information paths behind FPN), Balanced FPN \cite{DBLP:conf/cvpr/PangCSFOL19}, PAFPN \cite{LiuQQSJ18}. }
	\label{tab:modelextention}
	\vspace{-1mm}	
\end{table}
\subsection{Implementation Details}
\subsubsection{Datasets and Evaluation Criteria}
We conduct experiments on the COCO \cite{LinMBHPRDZ14} and PASCAL VOC \cite{EveringhamGWWZ10} benchmarks. For COCO, the training is conducted on the 118k training images, and ablation studies are evaluated on the 5k \texttt{minival} images. We also report the results on the 20k images in \texttt{test-dev} for comparison with state-of-the-art (SOTA). For evaluation, we adopt the metrics from the COCO detection evaluation criteria, including the mean Average Precisions (mAP) across IoU thresholds ranging from 0.5 to 0.95 at different scales. For PASCAL VOC, training is performed on the union of VOC 2007 trainval and VOC 2012 trainval (10K images) and evaluation is performed on VOC 2007 test (4.9K images), mAP with an IoU threshold of 0.5 is used for evaluation.

\subsubsection{Super-net Training and Sub-net Searching Phase}
We consider a total of $N=5$ intermediate nodes for super-net training and optimal sub-net searching. We choose Faster R-CNN (ResNet50 \cite{HeZRS16}) as the baseline. During super-net training, we use input-size $800\times 500$ and sample $1/5$ images from training-set of COCO to further reduce the search cost. As for PASCAL VOC, the input-size is set to $384\times 384$. We use SGD optimizer with initial learning rate 0.02, momentum 0.9, and $10^{-4}$ as weight decay. We train super-net for 12 epochs with batch-size of 16. Edge importance weight $\gamma$ is initialized as 1, and hyper-parameter $\mu$ is $10^{-4}$. In total, the whole search phase is completed in 4 days (1 day for super-net training and 3 days for optimal sub-net search) using 1 V100 GPU.

\begin{table}[t]
	\vspace{-1mm}
\begin{centering}
	\small
	\tabcolsep 0.02in{\scriptsize{}}
	\begin{tabular}
	{c|ccccc}
		\hline\hline
		\textbf{Dataset} &\textbf{Method}   &\textbf{Search on} &\textbf{FLOPs}&\textbf{Params}& \textbf{mAP}\\  
	\hline
		\multirow{4}{*}{\shortstack{COCO}}
		&Basline \cite{LinDGHHB17} &-&207&41.5M  & 38.6 \\
		&Auto-FPN &VOC &-&31.3M   &38.9 \\
	    &Auto-FPN &COCO &260G&32.6M   &40.5 \\
	    &\textbf{OPA-FPN@64} &VOC &128G&\textbf{29.8M}&41.5\\
		&\textbf{OPA-FPN@64} &COCO &124G &29.8M&\textbf{41.6 }\\
\hline
		\multirow{2}{*}{\shortstack{VOC}}
		&Basline \cite{LinDGHHB17} &-&207G&41.2M & 79.7 \\
		&Auto-FPN &VOC &-&31.2M   &81.8\\
	    &Auto-FPN &COCO &256G&32.5M   &81.3 \\
	    &\textbf{OPA-FPN@64} &VOC &127G&\textbf{29.5 M}&\textbf{82.7}\\
		&\textbf{OPA-FPN@64} &COCO &124G&29.6M &82.5\\
		\hline
		\end{tabular}{\scriptsize\par}
\par\end{centering}		
\vspace{1mm}		
	\caption{
	Comparisons of model transferability between different datasets with Auto-FPN \cite{Xu_2019_ICCV}.
	}
	\label{tab:cros-valid}
\vspace{-1mm}

\end{table}

\subsubsection{Full Training Phase}
In this phase, we fully train the searched model.SGD is performed to train the full model with batch-size of 16. The initial learning rate is 0.02; $10^{-4}$ as weight decay; 0.9 as momentum. Single-scale training with input $1333\times 800 $ size is trained for 12 epochs, and the learning rate is decreased by 0.1 at epoch 8 and 11. While multi-scale training (pixel size={$400 \sim 1400$}) is trained for 24 epochs with learning rate decreased by 0.1 at epoch 16 and 22. We use single-scale training for ablation studies if not specified, and we compare with SOTA with multi-scale training.

\subsection{Results}
\subsubsection{Comparison with SOTA}
\textit{The searched optimal architecture by OPANAS}, denoted as \textbf{OPA-FPN}, is illustrated in Fig.~\ref{fig:super-net} (c). 
It is evaluated together with state-of-the-art detectors
including hand-crafted \cite{DBLP:journals/corr/abs-1911-09070} and NAS-based ones \cite{DBLP:journals/corr/abs-1911-09929,  DBLP:conf/cvpr/GhiasiLL19,  DBLP:conf/cvpr/WangGCWTSZ20, DBLP:conf/cvpr/DuLJGTCLS20,  Xu_2019_ICCV, DBLP:conf/nips/ChenYZMXS19}.
Note that these methods search different components of detectors. Specifically, SM-NAS searches the overall architecture of Cascade R-CNN, NAS-FPN searches 
the neck architecture, and Auto-FPN searches  
the architectures of both neck (FPN) and detection head. 
As shown in Tab.~\ref{tab:comparenas}, compared with representative results achieved by these SOTA methods, our method achieves better or very competitive results in terms of amount of parameters, computation complexity, accuracy, and inference speed. 
Notably, our method can search for the best neck architecture more efficiently, \textit{e.g.}, 4 GPU days on COCO. 
Specially, our searched OPA-FPN equipped with Cascade R-CNN
Res2Net101-DCN \cite{8821313} achieves a new state-of-the-art accuracy-speed trade-off (52.2 $\%$ mAP at 7.6 FPS), outperforming  SpineNet (the SOTA NAS based method) and EfficientDet (based on the SOTA NAS searched backbone). These results demonstrate the effectiveness of our carefully designed search space and the efficiency of the search algorithm.
\begin{table}[t]
	\vspace{-1mm}
\begin{centering}
	\small
	\tabcolsep 0.06in{\scriptsize{}}

		\begin{tabular}{l|ccc}
			\hline\hline \textbf{Method}  &\textbf{FLOPs} &\textbf{Params} & \textbf{mAP}\\  
			\hline
			Baseline \cite{LinDGHHB17} &207G &41.5M  &36.4\\
	  	    Top-down &$209^{+0.7\%}$G &$41.8^{+0.5\%}$M  &$37.5^{+1.1}$\\
			Bottom-up &$209^{+0.7\%}$G &$41.8^{+0.5\%}$M  &$30.0^{-6.4}$\\
			Scale-equalizing &$234^{+13.0\%}$G &$41.1^{-0.9\%}$M  &$32.6^{-3.8}$ \\
		    Fusing-splitting &$\mathbf{181^{-12.4\%}}$G &$40.1^{-3.4\%}$M  &$37.6^{+1.2}$ \\
            \hline
			\textbf{OPA-FPN@112} &$197^{-3.4\%}$G &$\mathbf{35.5^{-14.5\%}}$M   &$\mathbf{39.6^{+3.2}}$\\
			\hline
		\end{tabular}{\scriptsize\par}
\par\end{centering}		
\vspace{1mm}		
	\caption{
		Comparisons with the single-information-path FPN architectures on COCO \texttt{minival}.} 
	\label{tab:search result}
	\vspace{-1mm}
\end{table}
\begin{figure}
	\centering
	\begin{minipage}{\linewidth}
		\centering
		\subfloat{{\includegraphics[width=.44\linewidth]{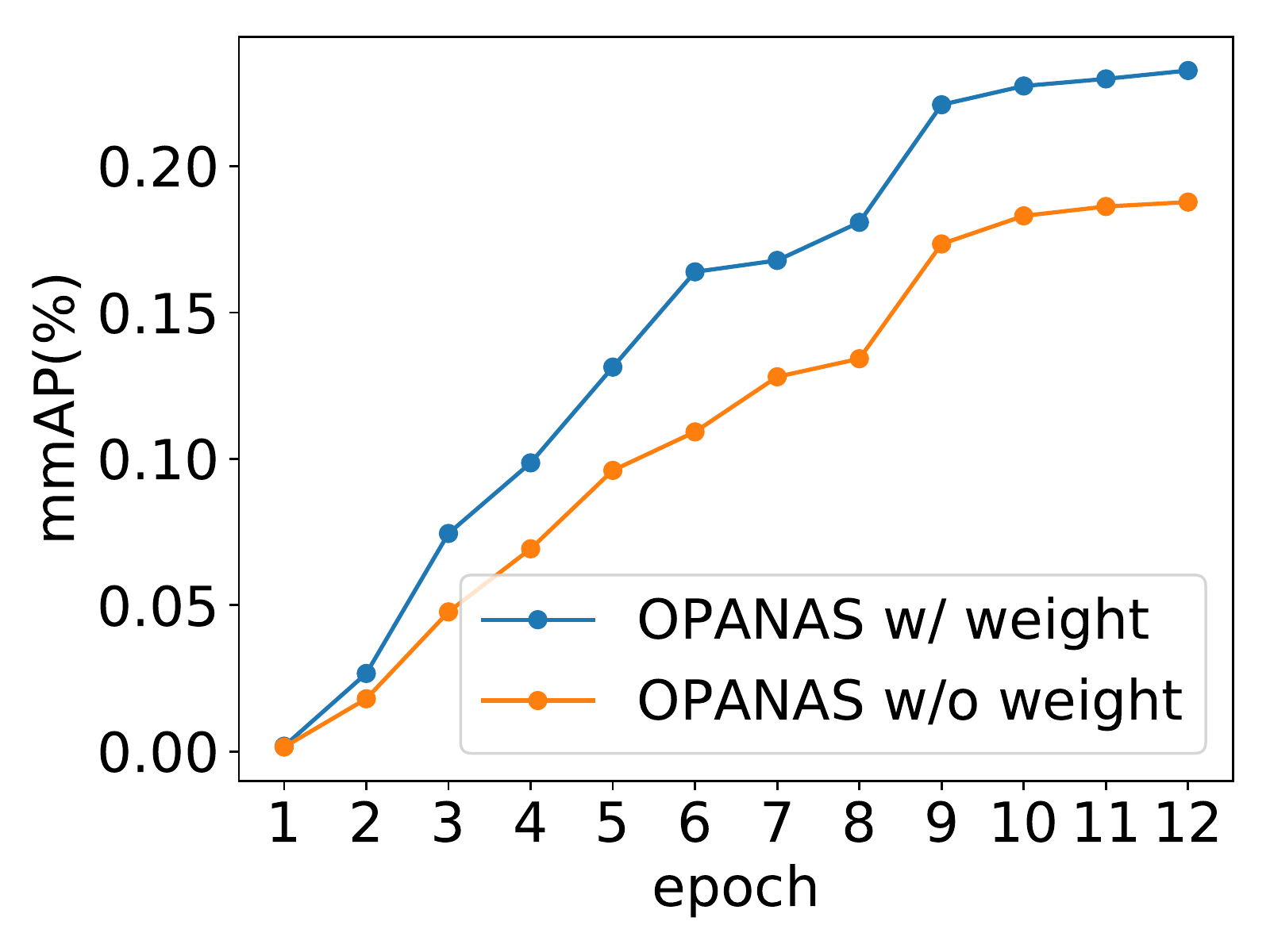} }}
		\qquad \subfloat{{\includegraphics[width=.44\linewidth]{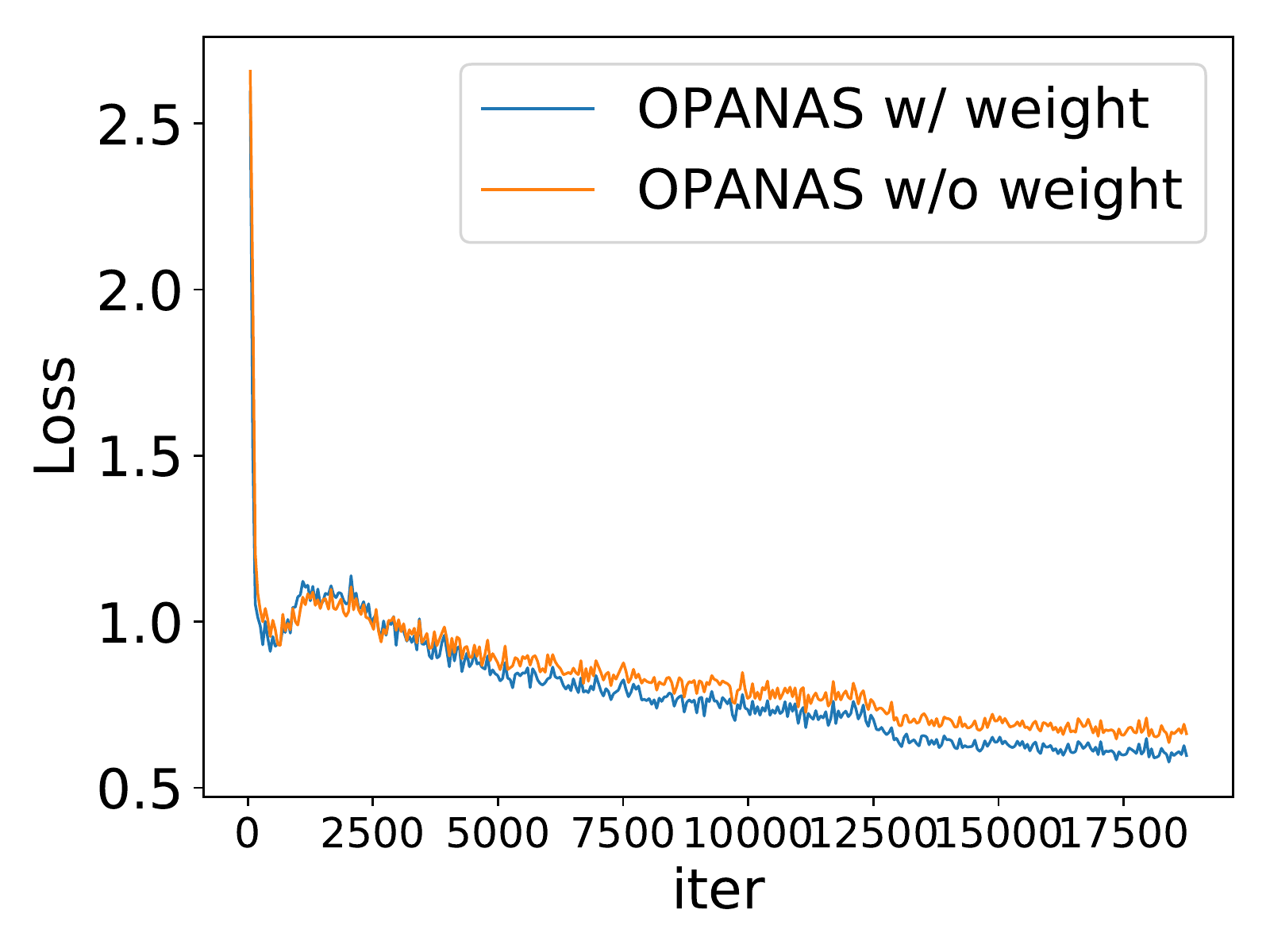} }}
		\qquad
				\vspace{-1em}
		\caption{ Comparisons of the intermediate results of super-net training w/ or w/o edge importance weighting. mmAP denotes the mean mAP of random sampled 50 sub-nets from the current super-net training epoch.}
		\label{fig:additional constraint}
	\end{minipage}
		\vspace{-1em}

\end{figure}
\subsubsection{Model Adaptability for Main-stream Detectors}
To further verify the performance of the OPA-FPN on main-stream detectors, we adapt it to RetinaNet \cite{LinGGHD17}, Faster R-CNN \cite{RenHGS15} and Cascade R-CNN \cite{FelzenszwalbGM10} in Tab.~\ref{tab:modelextention}. Under the same training strategy with baseline, we obtain a lighter model {with better performance on each detector: improving RetinaNet by 2.3$\%$ mAP with 13$\%$ FLOPs decreasing, and improving Cascade R-CNN by 2.5$\%$ mAP with 27$\%$ parameter amount decreasing}. Moreover, comparing with other hand-craft FPNs, our architecture achieves clearly better results in terms of amount of parameters, computation complexity, accuracy.

\subsubsection{Model Transferability Between COCO and VOC}
To evaluate the transferability of our architecture on different datasets, we transfer the searched architecture between COCO and VOC with multi-scale training, as shown in Tab.~\ref{tab:cros-valid}. With the architecture searched on COCO, our method boosts the performance by $3.0\%$ mAP for COCO, $2.8\%$ mAP for VOC, respectively. When searched on VOC, our method boosts the performance by $3.0\%$ mAP for VOC, $2.9\%$ mAP for COCO, respectively. No matter which dataset we search on, our architecture performs better than Auto-FPN \cite{Xu_2019_ICCV} (\textit{e.g.}, $41.6\%$ vs.$40.5\%$ in terms of mAP) with fewer computation cost (\textit{e.g.}, FLOPs 128G vs. 260G). These results further demonstrate the effectiveness of our method.

\begin{table}[t]
	\vspace{-1mm}
\begin{centering}
	\small
	\tabcolsep 0.06in{\scriptsize{}}

		\begin{tabular}
			{c|c|c|c|c}
			\hline\hline
			\textbf{Densely}  &\textbf{Fair}&\textbf{Edge Importance} &\multirow{2}{*}{$\mathbf{\tau}$}&\multirow{2}{*}{\textbf{mAP}}\\
\textbf{Connected}&\textbf{Sampling}&\textbf{Weight}&&\\  
			\hline
			&&&0.4390&38.4\\
			\ding{51}&&&0.3584&39.2\\
			\ding{51}&\ding{51}&&0.5865&39.5\\
			\ding{51}&\ding{51}&\ding{51}&\textbf{0.6145}&\textbf{39.6}\\
			\hline
		\end{tabular}
		{\scriptsize\par}
\par\end{centering}		
\vspace{1mm}	
	\caption{
		Correlation analysis of proposed method.} 
	\label{tab:correlation}
	\vspace{-1mm}
\end{table}

\subsection{Ablation Study}
\subsubsection{Information Path Aggregation}
To demonstrate the effectiveness of aggregating different information paths in our searched OPA-FPN,
we first compare it with the architectures using single information path in Tab.~\ref{tab:search result}. We choose the original Faster R-CNN (ResNet50) + vanilla FPN \cite{LinDGHHB17}
as the baseline, and we adjust the channel dimensions of our searched OPA-FPN and detection head, aiming to align the complexity with the baseline.
OPA-FPN significantly surpasses the baseline and the single-information-path architectures, with fewer FLOPs/Params (\textit{e.g,} FLOPs 197G vs. 207G, Params 35.5M vs. 41.5M), achieving an effective aggregation and exploration of information paths.

\subsubsection{Edge Importance Weighting}
To verify the effectiveness of the edge importance weight $\gamma$ in Eq. (\ref{eq:gamma}), we illustrate the intermediate results of training super-net with fair sampling in Fig.~\ref{fig:additional constraint}~(a,b), and observe that the edge importance weighting brings clear benefits for the training of super-net. These results prove that distinguishing the importance of different edges is effective for densely connected super-net training.

\subsubsection{Correlation Analysis}
Recently, the effectiveness of weight sharing-based NAS methods is questioned because of the lack of (1) 
fair comparison on the same search space and (2) adequate analysis on the correlation between the super-net performance and the stand-alone sub-net model performance \cite{DBLP:journals/corr/abs-1904-00420}. Here we adopt Kendall Tau~\cite{correlation} to measure the correlation of model ranking obtained from super-net. Specifically, we randomly sample 15 sub-nets from the trained super-net and conduct full train to evaluate their performance. In Tab.~\ref{tab:correlation}, when changing from single-path super-net used by SPOS \cite{DBLP:journals/corr/abs-1904-00420} to our densely connected super-net, there is a drop in correlation but the detection accuracy increases. 
However, by adopting fair sampling and edge importance weight, we achieve a much higher correlation value, showing that our method can achieve a higher correlation between super-net and sub-net with the fair sampling and the proposed edge importance weighting. 
\subsubsection{Comparisons with More NAS Baselines}
We further compare our method with more existing NAS methods, including a) Random Search: we randomly sample 15 architectures from the proposed search space and conduct full training under the same training setting in our experiments; b) SPOS: we train the single-path one-shot FPN super-net and perform EA search strategy following \cite{DBLP:journals/corr/abs-1904-00420}; c) DARTS: a very popular differentiable NAS method \cite{DBLP:conf/iclr/LiuSY19}; and d) Fair DARTS: an improved version of DARTS with softmax relaxation and zero-one loss \cite{DBLP:conf/cvpr/JiangXZLL20}. As the results reported in Tab.~\ref{tab:nasabla}, compared with other NAS methods, our method can find a better architecture with less or comparable time. These results prove the effectiveness of our super-net training and optimal sub-net search strategy.

\begin{table}[t]
	\vspace{-1mm}
	\small
\begin{centering}
	\small
	\tabcolsep 0.02in{\scriptsize{}}
	\begin{tabular}
	{p{35pt}|p{52pt}<{\centering}ccp{55pt}<{\centering}}	
		\hline\hline
		 \textbf{Search Method}  
		&\textbf{Search time (GPU days)} &\textbf{FLOPs} &\textbf{Params}&\textbf{Average/Best mAP of searched arch}\\  
		\hline
				
	Random  &55&231/206G &36.4/35.9M&37.5/39.1  \\
	SPOS  &4&207G &36.3M&38.4  \\
DARTS   &5&198G&37.8M&39.1\\
FairDarts   &4&269G &36.2M&39.4\\
OPANAS&4 &\textbf{197G} &\textbf{35.5M}   &\textbf{39.6}\\
		\hline
	\end{tabular}
{\scriptsize\par}
\par\end{centering}	
\vspace{1mm}
	\caption{
Comparisons with more NAS baselines on COCO \texttt{minival}. The baseline detector is Faster R-CNN ResNet50, the search space of SPOS is illustrated in Fig. \ref{fig:super-net}.1, and the others all adopt the proposed search space illustrated in Fig. \ref{fig:super-net}.2}
	\label{tab:nasabla}
	\vspace{-1mm}
\end{table}

\section{Conclusion}
In this paper, we propose One-Shot Path Aggregation Network Architecture Search (OPANAS), which consists of a novel search space and an efficient searching algorithm, to automatically find an effective FPN architecture for visual object detection. In particular, we introduce six types (\textit{i.e.}, top-down, bottom-up and scale-equalizing, fusing-splitting, balanced, skip-connect, and none) of information paths as candidate operations, and exploit densely connected DAG to represent FPN to aggregate them. An efficient one-shot search method is further invented to search the optimal FPN, based on the super-net training with fair sampling and edge importance weighting. Extensive experimental results demonstrate the superiority of the proposed OPANAS in both efficiency and effectiveness.

{\small
\bibliographystyle{ieee_fullname}
\bibliography{detection}
}

\end{document}